\documentclass[letterpaper, 10 pt, conference]{ieeeconf}
\IEEEoverridecommandlockouts
\usepackage[utf8]{inputenc}
\usepackage{graphicx}
\usepackage{multicol}
\usepackage{multirow}
\usepackage{array}
\usepackage{footmisc}
\usepackage{amssymb}
\usepackage{amsmath}
\usepackage{amsfonts}
\usepackage[linesnumbered,ruled,noend]{algorithm2e}
\usepackage[noend]{algpseudocode}
\usepackage[inkscapearea=page]{svg}
\usepackage{algpseudocode}
\usepackage{tikz}
\usepackage{float}
\usetikzlibrary{math,shapes}
\usepackage{tabularx}
\usepackage{booktabs}
\usepackage{pbox}
\usepackage{ragged2e}
\usepackage{multirow}
\usepackage{balance}
\usepackage{subfig}
\usepackage{relsize}
\usepackage{enumerate}
\usepackage{hyperref}
\usepackage{listings}
\mathchardef\mhyphen="2D
\DeclareMathOperator*{\argmin}{\arg\!\min}

\def\algo{Push-MOG}{}

\usepackage{etoolbox}
\makeatletter
\patchcmd{\@makecaption}
  {\scshape}
  {}
  {}
  {}
\makeatletter
\patchcmd{\@makecaption}
  {\\}
  { -\ }
  {}
  {}
\makeatother
\usepackage{siunitx}
\makeatletter
\renewcommand{\fnum@figure}{Figure~\thefigure}
\makeatother

\title{\LARGE \bf \algo: Efficient Pushing to Consolidate \\ Polygonal Objects
 for Multi-Object Grasping}
\author{Shrey Aeron$^1$, Edith Llontop$^1$, Aviv Adler$^1$, Wisdom C. Agboh$^{1,2}$, Mehmet Dogar$^2$, and Ken Goldberg$^1$
\thanks{$^1$The AUTOLab at UC Berkeley (automation.berkeley.edu)
{\tt\small \{aeron, goldberg\} @berkeley.edu}}
\thanks{$^2$University of Leeds, UK}
}
\SetKwRepeat{Do}{do}{while}
\begin{document}

\maketitle
\thispagestyle{empty}
\pagestyle{empty}

\begin{abstract}
Recently, robots have seen rapidly increasing use in homes and warehouses to declutter by collecting objects from a planar surface and placing them into a container. While current techniques grasp objects individually, Multi-Object Grasping (MOG) can improve efficiency by increasing the average number of objects grasped per trip (OpT). However, grasping multiple objects requires the objects to be aligned and in close proximity. In this work, we propose \emph{Push-MOG}, an algorithm that computes ``fork pushing'' actions using a parallel-jaw gripper to create graspable object clusters. In physical decluttering experiments, we find that Push-MOG enables multi-object grasps, increasing the average OpT by $\boldsymbol{34\%}$. Code and videos are available at \url{https://sites.google.com/berkeley.edu/push-mog}.
\end{abstract}

\section{Introduction}
Efficient object manipulation is a key task in industrial, commercial, and domestic robotics, incorporating elements from motion planning, grasping, and task planning
\cite{GOMP, DJ-GOMP, Herbert-CDC-2017, Ichnowski-ICRA-2022}. 
In particular, many applications focus on the task of transferring a collection of objects from a surface into a bin or basket, for the purpose of either clearing the surface or preparing the objects for transportation to another location. 
This task is typically solved by picking one object at a time.
However, an alternative approach of removing multiple objects at once with \emph{multi-object grasping} (MOG) has received relatively little attention. MOG can provide better efficiency than traditional single-object grasping \cite{mogplane, mognet}, especially when the bin is relatively far away.
Prior work on MOG mainly focused on developing techniques (such as MOG-Net \cite{mognet}) for detecting and executing effective multi-object grasps in a scene. However, grasping multiple objects at once requires all of them to be close enough to fit within the gripper width, which may be rare when the objects are distributed randomly.
Thus, to create graspable object clusters, pushing \cite{Arruda-Humanoids-2017, Hogan2020} can be useful.  

In this work, we propose \emph{\algo}, an algorithm that uses pushes to increase the average number of objects transported per trip to the bin.
\algo~uses hierarchical clustering to identify clusters of objects which could potentially be grasped together. To execute stable pushes on polygonal objects with a parallel-jaw gripper, \algo~utilizes \emph{fork pushing}, in which the jaw is opened to an appropriate width and the object is pushed perpendicular to the jaw, thus enabling a variety of stable pushes (see Fig.~\ref{fig:polypush}).

\begin{figure}[t]
    \vspace{0.2cm}
    \centering
     {\includegraphics[width=85.5mm]{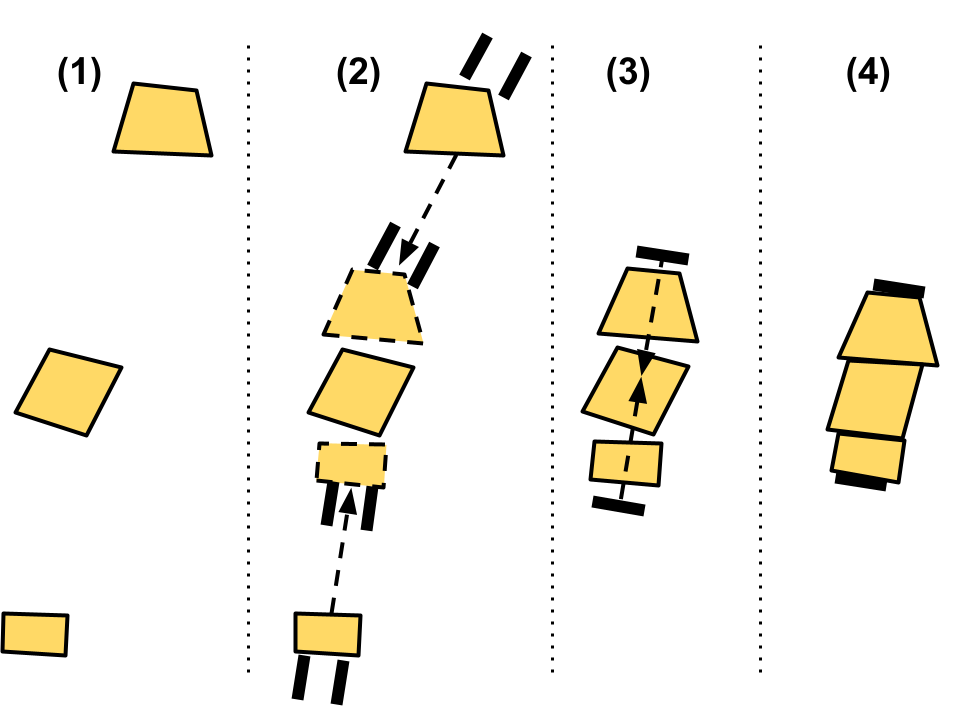}}
   \caption{An example of Push-MOG pushing and grasping objects into a cluster. (1) A cluster is identified containing three objects; (2) two objects are moved toward the center object using fork pushing. (3) Once the cluster fits within the robot's gripper, (4) the gripper closes to create a tight fit and grasps the objects that will be transported to the bin.}
   \vspace{-7mm}
\label{fig:push_alg}
\end{figure}

We also propose (mean) \emph{objects per trip} (OpT) as a metric for evaluating the effectiveness of multi-object grasps. Thus, in single-object grasping, the OpT is $1$ (or slightly less than $1$ due to failed grasp attempts), whereas MOG improves efficiency by increasing OpT. %OpT d. 
\label{sec:baselines_intro}
We evaluate \algo~in a physical environment consisting of randomly-distributed polygonal objects on a flat surface, in which the robot executes both grasps and pushes using a parallel-jaw gripper. We consider two baselines: (i) Single-Object Grasping (SOG), which removes objects at random, one at a time; (ii) and MOG-Net, which searches for multi-object grasps but does not apply rearrange objects with pushes. In physical robot experiments, we find that prior work (MOG-Net) achieves an OpT of $\approx 1$, suggesting that naturally occurring multi-object grasps are rare, while Push-MOG achieves an OpT of $1.339$, though at a cost of extra time for executing pushes. Since increasing OpT is more valuable when the bin is further away, \algo~can improve picking efficiency when trips are long.
We make the same assumptions as Agboh et al.~\cite{mogplane}: that the objects are extruded convex polygons, and have roughly uniform mass so that their centers of mass coincide with their geometric centers. However, unlike \cite{mogplane}, we also assume the objects are stable and do not topple when pushed.

In this work, we make the following contributions:
\begin{itemize}
    \item We formulate a new variant of the decluttering problem, in which a robot uses pushing actions to increase the number of objects it can move simultaneously to the bin. To solve this problem, we propose and implement \algo. We also propose objects per trip (OpT) as a metric for evaluating the effectiveness of MOG algorithms.
    \item We propose and implement \emph{fork pushing} as a way to quickly and stably move objects on a work surface using a parallel-jaw gripper.
    \item We evaluate the performance of \algo~and two baselines (one using only single-object grasping and one using multi-object grasping without pushes) and find that \algo~increases OpT by $34\%$.
\end{itemize}
\section{Related Work} 

\subsection{Multi-object grasps} 

Prior work on multi-object grasping \cite{Harada-ICRA-1998, Harada-TRA-2000, Yamada-ICRA-2005, Yamada-JCSE-2015} used multi-fingered robot hands and proposed conditions for a stable grasp of objects, 
focusing on numerical simulations without physical robot multi-object grasps or a focus on automation. 
Chen et al. \cite{Chen-IROS-2021} propose a method for a robot to dip its gripper inside a pile of identical spherical objects, close it, and estimate the number of grasped objects. Shenoy et al.~\cite{Shenoy-CoRR-2021} focus on the same problem but with the goal of transporting the picked spherical objects to another bin. 

Another line of work has focused on designing appropriate grippers for picking up multiple objects at once. Jiang et. al. \cite{jiang2023multipleobject} proposed a multiple suction cup vacuum gripper, while Nguyen et. al. \cite{Nguyen-MOG} proposed a soft gripper based on elastic wires for multi-object grasping. 

Our work is focused on efficiently rearranging scenes with pushes~\cite{Dogar-RSS-2011, Danielczuk-CASE-2018, Huang-IROS-2021, Agboh-Humanoids-2018} before multi-object grasping. Sakamoto et al. \cite{Sakamoto-IROS-2021} proposed such a picking system where the robot first uses pushing~\cite{largescalerearrange, Bejjani-IROS-2021, Mohammed-ICRA-2020}, to move one cuboid to the other and thereafter grasp both cuboids. Our approach handles any multi-sided polygonal object and focuses on large-scale problems where clustering is required to decide which objects to push and where. 

\subsection{Pushing} 
Pushing is a fundamental primitive in robot manipulation \cite{mason-book}.
Robotic pushing has been used to move a target object to a goal location \cite{Hogan2020, Agboh-ISRR-2019}, rearrange a working surface \cite{king2016, Song-IROS-2020}, and retrieve items from cluttered shelves \cite{Huang-IROS-2021, Agboh-Humanoids-2018, Muhayyudin-RAL-2018}.

During robotic pushing, uncertainty in the state, control, and model can result in failures~\cite{Agboh-WAFR-2018, Agboh-arxiv-2021}. Prior work \cite{Yuan-ICRA-2018, Arruda-Humanoids-2017} has trained networks to generate robust pushing actions. Others have taken an open-loop approach to generate these robust pushes \cite{Goldberg-Algorithmica-1993, Erdmann1988}. For example, Johnson et. al. \cite{Johnson-RAL-2016} propose robustness metrics and use them to generate robust open-loop pushes. They exploit gripper and object geometries to generate these robust pushes. We take a similar open-loop pushing approach and leverage object and gripper geometries to generate robust pushes. 

\subsection{Point Clustering}\label{sec:usup_clust}
Prior work in point clustering is well-established \cite{James2013, mirkin2005clustering}, specifically in the hierarchical variant \cite{heir, heirfast}, which aims to cluster data in an unsupervised fashion. This method constructs a distance-based tree that forms clusters of points from the bottom up which is then split at a `height' to determine the final clusters. 
Additional work by Dao et al. \cite{dao_constrain} explores clustering with human-defined constraints. It introduces the idea of simplifying and solving NP-Hard problems by constraining them through a limit on inter-cluster distance metrics, such as maximum cluster diameter, or within-cluster variation \cite{dao_constrain}. In this work, we use a distance-based hierarchical clustering algorithm but split object clusters to satisfy gripper width constraints. 
\section{Problem Statement} \label{sec:problem-statement}

Agboh et al \cite{mogplane} studied the problem of using an overhead camera and a parallel-jaw gripper to transport a collection $O$ of extruded polygonal convex objects (prisms) from a flat work surface to a bin or box adjacent to the workspace by taking advantage of \emph{multi-object grasping}, in which multiple objects are grasped simultaneously and transported together to the bin. 

In this work, we extend this problem by adding the \emph{fork push} action, in which the gripper pushes an object. By doing so, the robot can arrange the objects to create more efficient grasps. Since pushing is faster to execute than grasping and the objects are typically closer to each other than to the bin, using pushes to facilitate multi-object grasps can reduce the number of trips needed to clear the objects. As in Agboh et al. \cite{mognet}, we consider a frictional model of planar grasping, which not only corresponds better to real grasping problems but also permits a larger set of stable grasps, thus allowing the algorithm to consider a wider range of potential solutions.

\subsection{State, Action, and Objective}

Let the set of objects on the work surface be $O = \{o_0, o_1, \dots, o_{N_{o - 1}}\}$ where each $o_i$ is a convex polygonal object and $N_o$ is the number of objects on the work surface. 
The robot executes a sequence of \emph{pushes} and transports ( which we call \emph{trips}); each push translates and/or rotates a single object in the workspace without colliding with others, while each trip grasps a set of closely clustered objects and transfers them to the bin. 

We divide this problem into three sub-problems:
\begin{enumerate}
    \item \emph{Clustering:} Divide the objects into clusters. No cluster should contain more objects than can be stably grasped by the gripper.
    \item \emph{Pushing:} Push the objects in each cluster close together.
    \item \emph{Trip:} Perform grasps (containing as many objects as possible each time) and transport the grasped objects to the bin until no clusters remain.
\end{enumerate}
Due to the inherent uncertainty of working with real objects, a grasp might not get every object in a cluster, so a cluster may need multiple grasps to clear completely. The clustering step is purely computational and does not involve any physical action.

Since we don't optimize an explicit objective in the clustering or pushing problems, we instead aim to guide the choice of algorithm to maximize overall OpT.

\subsection{The Clustering Problem} \label{sec:clustering-problem}

In the clustering stage, the goal is to partition the set of objects $O$ into clusters in such a way that (i) the objects in any cluster are close together and (ii) each cluster consists of objects that can be stably grasped together (provided they are pushed into an appropriate configuration). Finally, since the goal of clustering the objects is to reduce the number of grasps necessary to clear the workspace, a good overall clustering will partition the objects into as few clusters as possible, as each cluster roughly corresponds to one grasp. We denote the output of the clustering algorithm as a list of clusters $C$, and each individual cluster as $c \subseteq O$ (since a cluster is a set of objects).

In this problem, each object $o$'s position is represented by its geometric centroid, which we denote $M_o$. This allows the application of an appropriately modified version of the hierarchical clustering algorithm (which clusters points).

An important characteristic of each object $o$ is its \emph{minimum final grasp diameter} $d^*_{o}$, corresponding to the minimum width of a stable single-object grasp on $o$. Then, given a cluster $c$, we denote the \emph{minimum grasp diameter of $c$} as $d^*_c = \sum_{o \in c} d^*_{o}$. The gripper also has a width $d^{\text{(gr)}}$, and if $d^*_c > d^{\text{(gr)}}$ we regard cluster $c$ as ungraspable (thus requiring division into smaller clusters)\footnote{It is possible that even if $d^*_c \leq d^{\text{(gr)}}$ the cluster does not have a stable grasp. For example, a cluster of three identical equilateral triangles cannot be stably grasped at all.}. 

Also, it is important for the clusters to not interfere with each other during the pushing step, which might happen if clusters are spatially intermingled. While point clustering usually avoids this (such clusters are locally non-optimal), the grasp diameter constraint may make such a solution to be `optimal' unless it is specifically excluded. We thus add a constraint that for any $c_1, c_2 \in C$, the centroids of the objects in $c_1$ (i.e. the set $\{M_o : o \in c_1\}$) are linearly separable from the centroids of the objects in $c_2$.

The clustering problem is then defined to partition $O$ into a set of clusters $C$ such that: (1) $d^*_c \leq d^{(\text{gr})}$ for all $c \in C$; (2) any $c_1, c_2 \in C$ have a line which separates the centers of mass of the objects they contain; (3) each cluster has small pairwise distances between its objects; and (4) there are few clusters overall. It is solved using a modified hierarchical clustering algorithm.

\subsection{The Pushing and Grasping Problems}

After the clusters have been defined, the next step is to push each cluster closer together to make multi-object grasping more effective (see Fig.~\ref{fig:push_alg} for an illustration). We assume that (due to the separating line constraint in the clustering problem) each cluster can be treated as a separate instance of the pushing problem.

Finally, once the pushing is done, we move the objects to the bin via multi-object grasping. As in \cite{mognet}, the objective is to move the objects to the bin with as few grasps as possible (without pushing, as that has already been done).

\begin{figure}
    \centering
    \includegraphics[width=85mm]{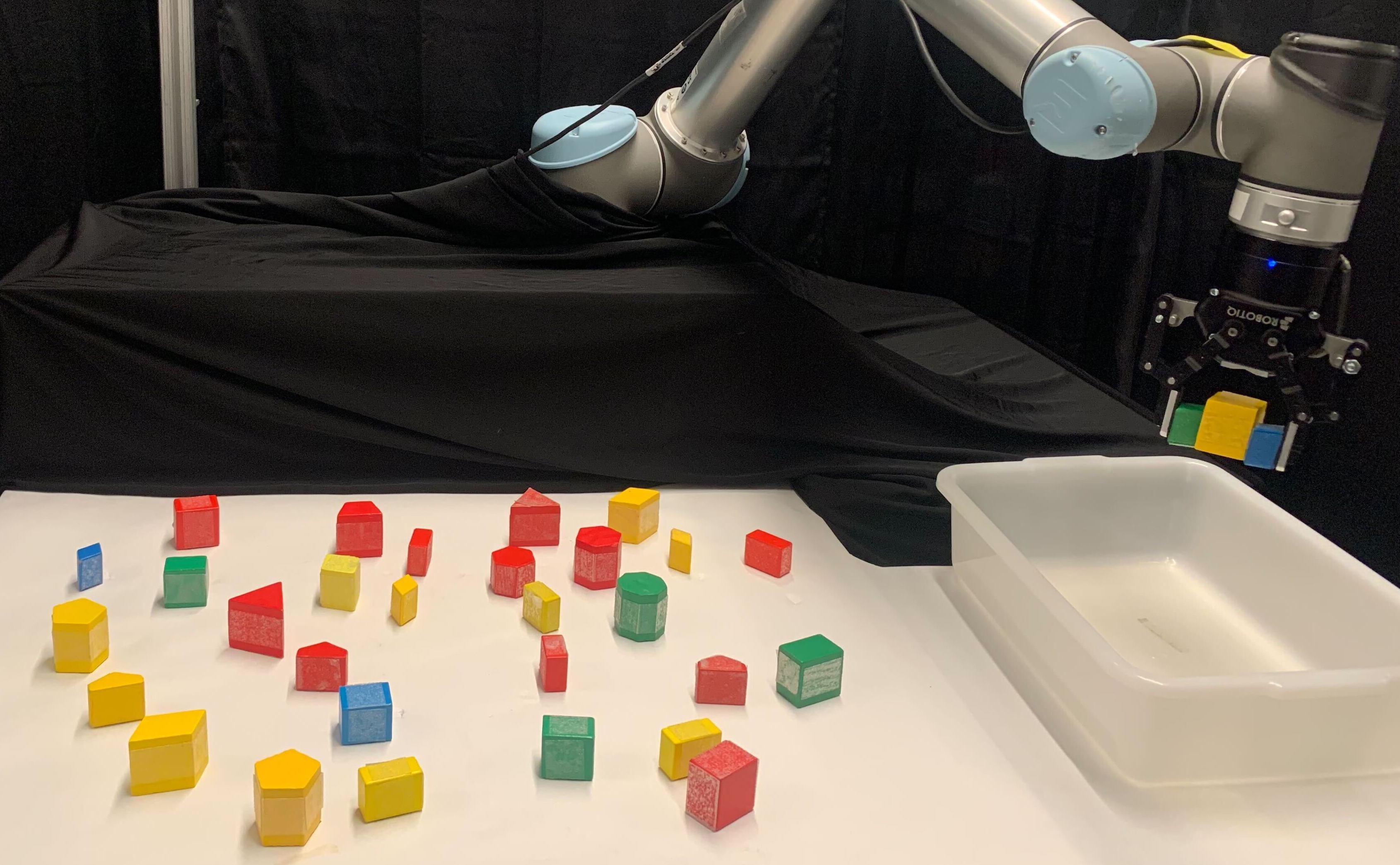}
    \caption{As illustrated in Fig.~\ref{fig:push_alg}, Push-MOG uses ``fork pushes'' to consolidate objects for multi-object grasping as shown on the right.}
    \label{fig:fig1}
    \vspace{5mm}
\end{figure}
\section{\algo{} Algorithm}
We present the \algo~algorithm in~Alg. \ref{alg:algo-overview}. After perceiving an initial scene, it computes a cluster of objects. We plan and execute a series of pushing actions to bring together the clustered objects so that they are touching each other. Finally, we query the MOG-Net algorithm \cite{mognet} to transport that cluster through Multi-Object grasps. Below, we outline how the \algo~algorithm plans and executes these steps.

\setlength{\textfloatsep}{2mm}
\begin{algorithm}[t]
    \SetKwInOut{Input}{Input}
    \SetKwInOut{Output}{Output}
    \SetKwInOut{Parameters}{Parameters}
    \SetKwInOut{Subroutines}{Subroutines}
    \Do{objects remain on the workspace}{
    $I$ $\gets$ Image of the workspace\\
    Parse $I$ into a 2D object map of all objects $O$\\
    Partition $O$ into clusters $C$ (Alg. 2, see \ref{sec:clustering})\\
    Get a cluster $c$ from $C$ at random\\
    Plan the pushes, $P$ (Alg. 3, see \ref{sec:push})\\
            \For{push $p \in P$}{
                Predict post-push location of pushed object
            }
        Run MOG-Net (\ref{sec:MOGnet}), using predicted post-push locations, to identify grasp  $g$ \\
        Execute $g$ to remove $c$\\
    }

    \caption{\algo{}}\label{alg:algo-overview}
\end{algorithm}
\setlength{\floatsep}{2mm}

\subsection{Clustering Details (Alg~\ref{alg:algo-cluster})}\label{sec:clustering}

As described in Section \ref{sec:clustering-problem}, the goal of this step is to cluster objects into graspable subsets. In order to accomplish this we compute the centroids of each object and use a bottom-up hierarchical clustering. We find the smallest bounding box that encloses each object and cluster them based on their proximity to each other. Once these clusters are formed, we check that the sum of the widths of all the objects in the cluster does not exceed the width of the gripper. If we encounter a non-valid cluster $c$ (i.e. such that $d^*_c > d^{(\text{gr})}$), then we split $c$ into two clusters $c_1, c_2$ using a separating line (objects are divided based on their centroids), minimizing the difference between the  cluster grasp diameters $|d^*_{c_1} - d^*_{c_2}|$.

\setlength{\textfloatsep}{2mm}
\begin{algorithm}[t]
    \SetKwInOut{Input}{Input}
    \SetKwInOut{Output}{Output}
    \SetKwInOut{Parameters}{Parameters}
    \SetKwInOut{Subroutines}{Subroutines}
    \Input{$O$: List of objects}
    \Output{$C$: List of clusters}

    Compute centroids $M_{o}$ for all $o \in O$ \\
    Hierarchical clustering on $\{M_{o} : o \in O \}$ to generate clusters $C$\\
    Check the validity of each cluster $c \in C$ based on max gripper width\\
    \For{non-valid cluster $c \in C$}{
        Minimize weight $W$ over $\theta$ using the following:
        \For{$\theta \in [-{\frac \pi 4}, {\frac \pi 4}]$}{
            $v = \langle\cos\theta,\sin\theta\rangle$\\
            Split the objects into two clusters $c_{1}, c_2$ on an infinite line defined by $v$, through the centroid of $c$ \\
            For this $\theta$, calculate weight difference $W = |\sum_{o\in c_1} d^*_o - \sum_{o\in c_2} d^*_o|$}
        Split the cluster on this $\theta$
    }
    \caption{Clustering Algorithm}\label{alg:algo-cluster}
\end{algorithm}
\setlength{\floatsep}{2mm}

\subsection{Push Planning (Alg~\ref{alg:algo-push})}\label{sec:push}

Push planning considers how to push objects in a cluster closer together in order to facilitate grasping. This can be broken into two steps: (i) determining the desired locations of the cluster's objects; (ii) computing the parameters of the pushes to get them there. For object $o$, we denote the desired location of its centroid as $M'_o$, which is chosen both to make pushing easier and to position the objects for grasping. For simplicity, we only specify a desired location for the centroid (and not a desired orientation). All pushes are executed in one motion in a straight line; the direction and distance that $o$ is pushed is given by the vector $M'_o - M_o$ (to move its center of mass from $M_o$ to $M'_o$).

\begin{figure}[t]
    \vspace{0.2cm}
    \centering
   {\includegraphics[width=85.5mm]
   {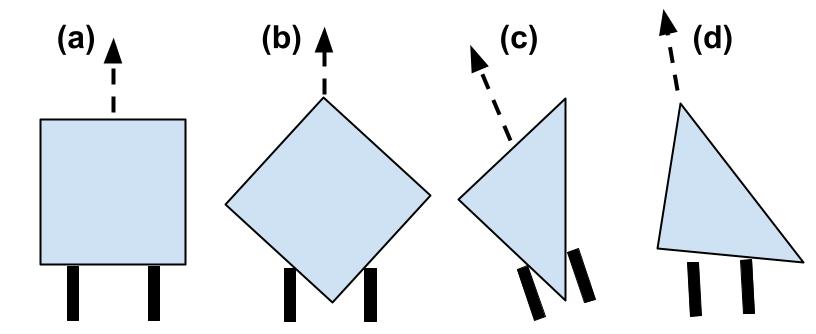}}
   \caption{Examples of fork pushing with a parallel-jaw gripper, both against an edge ((a) and (d)) and around a vertex ((b) and (c)) noticing that the direction the object will move to is roughly perpendicular to the parallel-jaw gripper.}
\label{fig:polypush}
\end{figure}

For executing a push with a parallel-jaw gripper, we propose the technique of \emph{fork pushing}: we angle the gripper so the line between the jaws is perpendicular to the desired push direction. This allows the gripper to push both along an edge or around a vertex of an object (see Fig.~\ref{fig:polypush}), whereas a flat-edged pusher could not push against a vertex without the object undergoing a major rotation, deviating from the desired path.

\setlength{\textfloatsep}{2mm}
\begin{algorithm}[t]
    \SetKwInOut{Input}{Input}
    \SetKwInOut{Output}{Output}
    \SetKwInOut{Parameters}{Parameters}
    \SetKwInOut{Subroutines}{Subroutines}
    \Input{$C$: List of Clusters}
    \Output{$v_{start}, v_{end}$: Start and end push points for each object}

    \For{valid cluster $c \in C$}{
        $M_c = \frac {\sum_{\text{object } o\in c} M_o}{|c|}$\\
        $m = \argmin_{o\in c} \|M_o - M_c\|$, the central object in the cluster\\
        \For{object $o \in c \backslash m$}{
            $v = M_o - M_m, v_l$ line with endpoints $M_o, M_m$\\
            Calculate center of furthest edge $e_o$ and corner $d_o$ of object $o$ from $m$, which intersects with $v_l$\\
            Calculate closest edge $e_m$ on $o$ to $o_m$ on $v_l$\\
            Scale $v \to v_p$ to capture the length of the push vector $\|v\|$\\
            Ensure no collisions by shortening $v_p$, such that $v_p + o$ does not collide with $o_m$
        }
    }
    \caption{Push Planning Algorithm}\label{alg:algo-push}
\end{algorithm}
\setlength{\floatsep}{2mm}

\subsection{MOG-Net Integration}\label{sec:MOGnet}
Following the pushing step, we use MOG-Net \cite{mognet} on the clusters to clear the workspace. To improve efficiency, we use the simulated coordinates of the push action to plan grasps instead of taking an image of the workspace again.

However, this process is not fully robust, because pushes may cause other blocks in the way to deviate from their estimated location, leading to a faulty grasp. 

To remedy this, whenever a trip is performed (thus moving the gripper out of the camera's view of the workspace), a new image of the workspace is taken, allowing these errors to be corrected; the algorithm also uses this image to re-plan the clustering (\ref{sec:clustering}) and pushing (\ref{sec:push}) on the remaining blocks (which may have been pushed aside or missed during an earlier grasp).

\section{Experiments and Results}\label{sec:experiments}

To evaluate the performance of the policies developed in the paper, we run experiments in a real workspace. 

\subsection{Experimental Setup}\label{sec:expsetup}

\begin{figure}
    {\includegraphics[width=85mm]{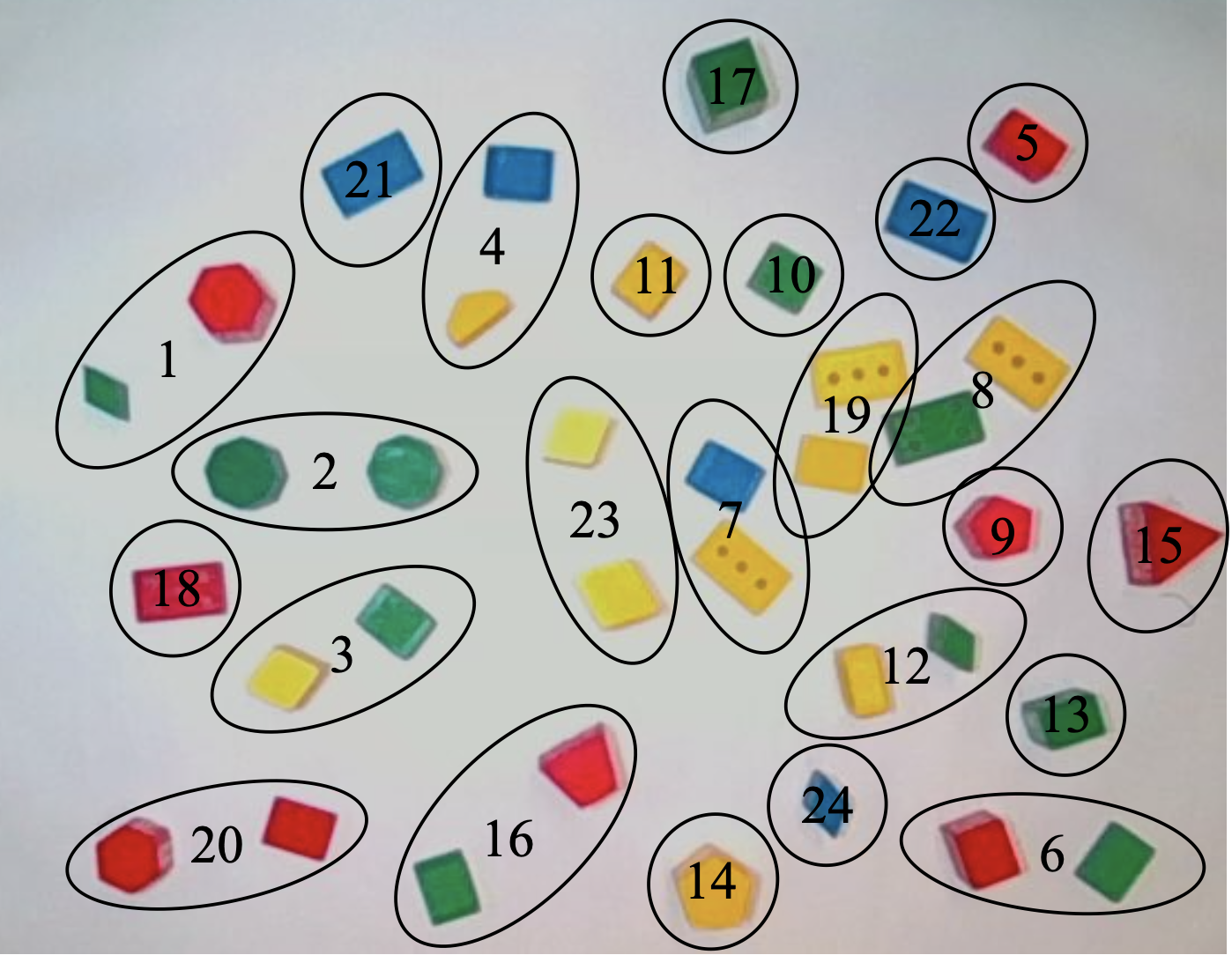}}
    \caption{This figure shows the clustering formed by Push-MOG of an initial scene. The numbers on each cluster identify the order that the algorithm will perform the pushing and grasping actions.}
    \label{fig:wksp_scatter}
\end{figure}

An example input is shown in Fig.~\ref{fig:wksp_scatter}. We use a Universal Robotics (UR) 5 Robot with a Robotiq 2F-85 gripper mounted to its wrist. We perform experiments on 34 convex objects of various sizes, ranging from 3-sided to 8-sided. All experiments assume a random placement of the objects in the workspace, and one such configuration is shown in Fig.~\ref{fig:wksp_scatter}. To generate a random placement, we uniformly sample the workspace and create circles that encompass the max radius of any object we are using. Before creating each circle we ensure that they do not overlap, and continue sampling the object set without replacement until all objects are placed. Then a human sets up the configuration in real, mirroring the randomly generated image. We generated 5 random scenes, on which we will perform all experiments. We add friction on the graspable faces of the objects by wrapping them in frictional tape. 

To perceive the environment, we use an Intel RealSense L515 Camera, which outputs RGB-D images, mounted directly above the workspace. To calibrate the Camera-Robot transform, use an ArUco marker \cite{aruco} which is manually calibrated with an accuracy of $\sim 1cm$.

The pushing action consists of 5 repeated steps:
\begin{enumerate}
    \item Open the jaws to 30 percent width
    \item Move the gripper above the desired push location
    \item Move down to the object's Z-height
    \item Move the gripper and push it to its planned location
    \item Move the gripper above its planned location
\end{enumerate}

The Multi-Object Grasping action is executed with the MOG-Net algorithm \cite{mognet}. Here we repeat the following:
\begin{enumerate}
    \item Plan a grasp for each cluster
    \item Move the gripper above the desired grasp location
    \item Open the jaws fully
    \item Move down to the work surface
    \item Close the jaws to complete the grasp
    \item Move the gripper outside the workspace and open the jaws fully to drop the objects into the basket
\end{enumerate}

\subsection{Baselines}

We compare Push-MOG against two baselines: Frictional SOG, which uses single-object grasps, and MOG-Net \cite{mognet}, which uses multi-object grasps but does not use additional actions to help group the objects together. See Section~\ref{sec:baselines_intro} for detailed information on baselines.

\subsection{Results}\label{sec:results}

Results are given in Table~\ref{tab:results}. We find that randomly-placed objects are not generally well-positioned for multi-object grasps, as OpT is almost identical between MOG-Net and Frictional SOG, thus requiring the use of additional actions such as pushes to assist MOG. Indeed, the similarity in the performance of Frictional SOG and MOG-Net suggests that there are few if any, `naturally occurring' multi-object grasps in a typical scene of randomly-scattered objects. \algo~successfully uses the pushing action to generate graspable object clusters, increasing OpT by roughly $34\%$. This comparatively modest increase in OpT (as compared to the experimental results on the original MOG-Net \cite{mognet}) is because objects that are stable under pushing tend to have larger minimum grasp diameters and the gripper we used has a relatively small max opening width, so it can grasp at most 2 or 3 objects at a time.
\begin{center}
\begin{table}[t]
\vspace{3mm}
\caption{Physical decluttering experimental results for random scenes, comparing baselines \textbf{Frictional SOG, MOG-Net}, and the method \textbf{\algo}. Experiments were conducted with a Robotiq 2F-85 parallel-jaw gripper mounted on a UR5 robot arm, on 5 random scenes of 34 objects, with the goal placed directly adjacent (see Figs \ref{fig:fig1} and  \ref{fig:wksp_scatter}).}
\centering 
\begin{tabular}{@{}l@{\quad} c@{\quad} c@{}}
\toprule 
Methods & Objects per Trip (OpT)\\
Frictional SOG & 0.993 $\pm$ 0.014 \\
MOG-Net & 0.986 $\pm$ 0.038 \\
\algo & 1.339 $\pm$   0.145\\
\bottomrule
\end{tabular}
\label{tab:results}
\end{table}
\end{center}

\section{Limitations and Future Work}
This work proposes \algo, a novel algorithm that consolidates polygonal objects into optimal clusters which increase the number of objects per grasp. This work has the following limitations: 1) pushing can cause misalignments where grasps fail 2) \algo~does not always avoid collisions 3) total time is slow, leading to fewer picks per hour, even though Push-MOG gets more objects per trip. In future work, we can incorporate \emph{vertical stacking} to create larger multi-object clusters. 
\section*{Acknowledgements}
\vspace{-0.1cm}
\relsize{-1}

This research was performed at the AUTOLAB at UC Berkeley in
affiliation with the Berkeley AI Research (BAIR) Lab, and the CITRIS ``People and Robots" (CPAR) Initiative. The authors were supported in part by donations from Toyota Research Institute, Bosch, Google, Siemens, and Autodesk and by equipment grants from PhotoNeo, NVidia, and Intuitive Surgical. Mehmet Dogar was partially supported by an EPSRC Fellowship (EP/V052659). We thank our colleagues who provided helpful feedback and suggestions, in particular Mallika Parulekar and Zehan Ma.

{\let\clearpage\relax \vspace{5mm} \bibliography{main}}

\end{document}